\documentclass[runningheads]{llncs}

\usepackage{eccv}

\usepackage{eccvabbrv}
\usepackage{multicol}
\usepackage{multirow}
\usepackage{caption}
\usepackage{subcaption}
\usepackage{graphicx}
\usepackage{booktabs}
\usepackage[accsupp]{axessibility}  

\usepackage{hyperref}

\usepackage{orcidlink}

\begin{document}

\title{A Graph-Based Approach for Category-Agnostic Pose Estimation}


\author{Or Hirschorn \and Shai Avidan}


\institute{Tel-Aviv University \\
\email{orhirschorn@mail.tau.ac.il} ~and~ \email{avidan@eng.tau.ac.il}\\
\url{https://orhir.github.io/pose-anything/}}

\maketitle


\begin{abstract}
Traditional 2D pose estimation models are limited by their category-specific design, making them suitable only for predefined object categories. This restriction becomes particularly challenging when dealing with novel objects due to the lack of relevant training data.
To address this limitation, category-agnostic pose estimation (CAPE) was introduced. CAPE aims to enable keypoint localization for arbitrary object categories using a few-shot single model, requiring minimal support images with annotated keypoints. 

We present a significant departure from conventional CAPE techniques, which treat keypoints as isolated entities, by treating the input pose data as a graph. We leverage the inherent geometrical relations between keypoints through a graph-based network to break symmetry, preserve structure, and better handle occlusions.
We validate our approach on the MP-100 benchmark, a comprehensive dataset comprising over 20,000 images spanning over 100 categories. Our solution boosts performance by 0.98\% under a 1-shot setting, achieving a new state-of-the-art for CAPE. Additionally, we enhance the dataset with skeleton annotations.
Our code and data are publicly available.
\setlength{\intextsep}{12pt}
\begin{center}
\begin{figure}[!h]
  \centering
    \setlength{\fboxsep}{0pt}
    \setlength{\fboxrule}{1.5pt}
    \setlength{\tabcolsep}{0.8pt}
    \renewcommand{\arraystretch}{0.8}
    \begin{tabular}{cccc}
    \multicolumn{2}{c}{\textbf{\textcolor{purple}{Support}}  $\rightarrow$  \textbf{\textcolor{ForestGreen}{Query}}} & \multicolumn{2}{c}{\textbf{\textcolor{purple}{Support}}  $\rightarrow$  \textbf{\textcolor{ForestGreen}{Query}}}\\
    
    \fcolorbox{purple}{white}{\includegraphics[width=0.22\textwidth]{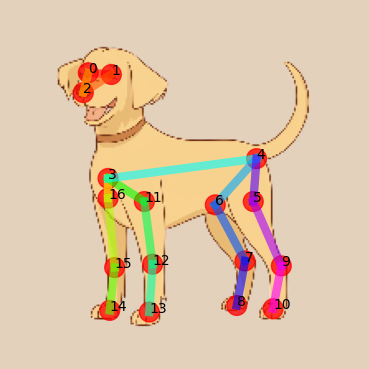}} &
    \fcolorbox{ForestGreen}{white}{\includegraphics[width=0.22\textwidth]{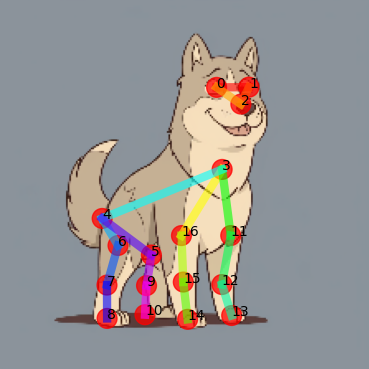}} &
    \fcolorbox{purple}{white}{\includegraphics[width=0.22\textwidth]{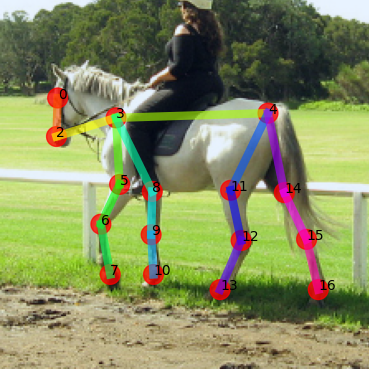}} &
    \fcolorbox{ForestGreen}{white}{\includegraphics[width=0.22\textwidth]{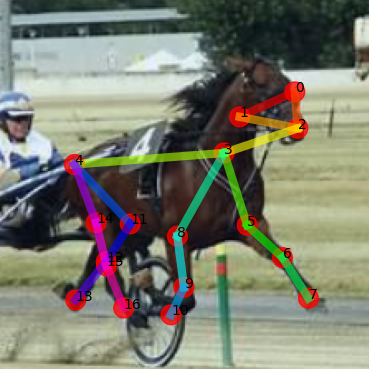}} \\
    
    \fcolorbox{purple}{white}{\includegraphics[width=0.22\textwidth]{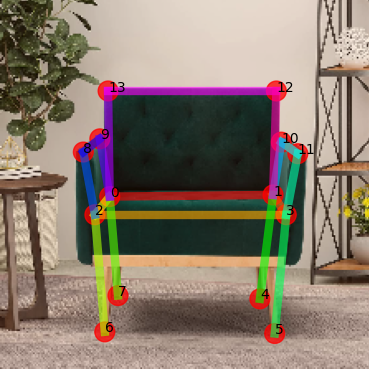}} &
    \fcolorbox{ForestGreen}{white}{\includegraphics[width=0.22\textwidth]{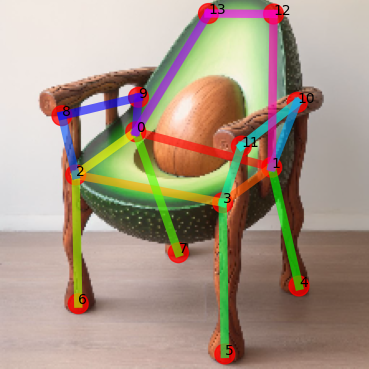}} &
    \fcolorbox{purple}{white}{\includegraphics[width=0.22\textwidth]{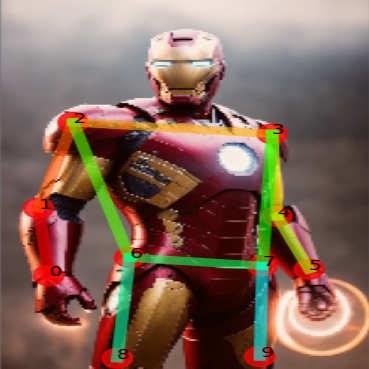}} &
    \fcolorbox{ForestGreen}{white}{\includegraphics[width=0.22\textwidth]{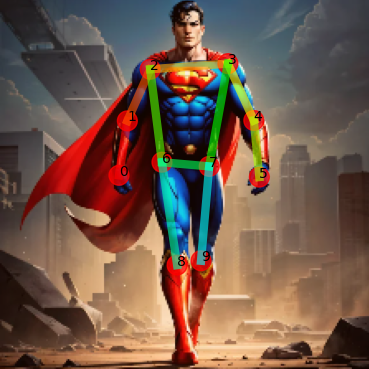}}  \\
    \end{tabular}
  \caption{Given a support image and skeleton from any category (\textcolor{purple}{purple}) our model localizes the skeleton on a query image (\textcolor{ForestGreen}{green}). Our graph-oriented method integrates structural information to improve keypoint localization. 
  }
  \label{fig:teaser}
\end{figure}
\end{center}
\end{abstract}
    
\section{Introduction}
\label{sec:intro}

2D pose estimation, also known as keypoint localization, has recently gained significant prominence in computer vision research, finding diverse applications in both academic and industrial domains. This task involves predicting specific semantic parts' locations within an object depicted in an image. Notably, it plays a crucial role in areas such as human pose estimation for video understanding or virtual reality, animal pose estimation in zoology, and vehicle pose estimation for autonomous driving.

However, a fundamental limitation of conventional pose estimation models is their inherent category specificity. These models are typically designed to work exclusively within a predefined object category, restricting their use to the domain they were trained on. Consequently, their adaptability to real-world situations involving novel objects is impeded due to the absence of relevant training data.

To address this challenge, category-agnostic pose estimation (CAPE) was proposed~\cite{xu2022pose}. CAPE seeks to perform keypoint localization for arbitrary object categories, using a single model, by utilizing just one or a few support images with keypoint annotations, referred to as support keypoints.
This approach enables the generation of an object's pose based on arbitrary keypoint definitions. Significantly, it substantially reduces the extensive costs associated with data collection, model training, and parameter tuning for each novel class. This paves the way for more versatile and adaptable applications in the field of pose estimation.
Figure~\ref{fig:teaser} shows some results of our work.



CAPE methods match support keypoints with their corresponding counterparts in the query image. The matching is done in latent space to facilitate the localization of keypoints. 
Previous CAPE methods treat keypoints as individual, disconnected points. However, inherent geometrical relations between keypoints serve as a robust prior that can enhance the accuracy of keypoint localization by breaking symmetry and handling occlusions.
In contrast to these approaches, we recognize this significant geometrical structure and leverage it by treating the input keypoints as a graph.
Building upon this realization, we introduce GraphCape, a graph-based model, purposefully designed to capture and incorporate this crucial structural information. By doing so, we exploit the inherent relationships and dependencies between keypoints.
We explored various design choices, detailed in the supplementary, and ultimately chose to incorporate this prior into the model architecture.

We evaluate our method on the category-agnostic pose estimation benchmark MP-100~\cite{xu2022pose}. 
This dataset contains over $20k$ images from over 100 categories, including animals, vehicles, furniture, and clothes. It is composed of samples collected from existing category-specific pose estimation datasets. As some of the skeleton data was missing, we collected skeleton annotations from the original datasets and annotated several categories with missing skeleton definitions. 

It is important to note that the required additional keypoints connectivity input is shared among instances of the same category. Thus, the additional graph annotations required equals the number of categories. For example, when using our method for fine-tuning or inference on a specific category, the annotation process requires adding only one skeletal definition, which is quite simple.

We compare our method to previous CAPE methods, as well as an updated version of the previous state-of-the-art which we call CapeFormer-T. Our method improves over this updated version, boosting performance under a 1-shot setting by 0.98\%, achieving a new state-of-the-art.

To summarize, we propose three key contributions:
\begin{itemize}
  \item We propose to treat the input keypoints as connected nodes of a graph, instead of independent entities. Furthermore, to leverage the graph connectivity we introduce a new graph-based model design.
  \item We provide an updated version of the MP-100 dataset with skeleton annotations for all categories.
  \item We achieve state-of-the-art performance on the MP-100 benchmark.
\end{itemize}

\section{Related Works}
\label{sec:related}
\subsection{Detection Transformer}
A compelling connection emerges between detection problems, focused on bounding box estimation and localization, and pose estimation, which centers on keypoint localization. Recognizing this connection, we incorporate several advancements from object detection models into the latest category-agnostic pose estimation method~\cite{Shi_2023_CVPR}, thus improving its performance.

The DEtection TRansformer (DETR)\cite{carion2020end} was the first transformer-based network for object detection, replacing conventional spatial anchors with learnable object queries in a transformer decoder. Its simplicity and universally applicable approach require minimal domain-specific knowledge, avoiding customized label assignments or non-maximum suppression. However, the original DETR design faced challenges such as slow convergence rates and reduced detection accuracy. In response, numerous studies\cite{meng2021CondDETR,gao2021fast,dai2021dynamic,zhu2020deformable,zhang2022dino,liu2022dab} have emerged to improve the DETR paradigm. These works have led to the development of top-tier object detectors, leveraging innovations like the reintroduction of multi-scale features and local cross-attention computations~\cite{zhang2022dino,fang2022eva,wang2022internimage}.

\subsection{Category-Agnostic Pose Estimation}

The primary goal of pose estimation is to localize the semantic keypoints of objects or instances accurately. Traditionally, pose estimation methods have predominantly been tailored to specific categories, such as humans~\cite{alphapose, openpose, yang2021transpose}, animals~\cite{yu2021ap, yang2022apt}, or vehicles~\cite{song2019apollocar3d, reddy2018carfusion}.
Existing research focused mainly on designing convolutional neural networks~\cite{alphapose, openpose, kreiss2021openpifpaf} or transformer-based architectures~\cite{mao2021tfpose, xu2022vitpose}. However, these methods are limited to categories encountered during training.

A relatively unexplored area is category-agnostic pose estimation (CAPE), introduced by Xu~\etal\cite{xu2022pose}. This approach predicts keypoints by comparing support keypoints with query images in the embedding space, handling unseen object categories during training. This shifts pose estimation into few-shot learning, which utilizes prototype, meta-learning, and fine-tuning.
ProtoNet~\cite{snell2017prototypical} learns a prototype for each class in the support data and classifies query data based on the nearest prototype. MAML~\cite{finn2017model} is a meta-learning-based approach that finds optimal initialization weights for rapid generalization to novel tasks with minimal fine-tuning. Another few-shot approach is fine-tuning~\cite{nakamura2019revisiting}, where the model is pre-trained on all base categories and fine-tuned on support images of novel categories during testing.

Specifically designed for CAPE, POMNet~\cite{xu2022pose} employs a transformer to encode query images and support keypoints, with a regression head predicting similarity directly from the concatenation of support keypoint and query image features. 
CapeFormer~\cite{Shi_2023_CVPR} extends the matching paradigm to a two-stage framework, rectifying unreliable matching outcomes to enhance prediction precision.
We first update CapeFormer, enhancing their methodology. Then, we focus on the importance of geometrical structure, integrating it into our model design.

\subsection{Graph Neural Networks in Computer Vision}
Originally designed for graph data like social networks~\cite{sage}, citation networks~\cite{sen2008collective}, and biochemical graphs~\cite{wale2008comparison}, GCNs have become pivotal in computer vision tasks~\cite{xu2017scene,landrieu2018large,wang2019learning,jing2022learning}, finding applications in scene graph generation, point cloud classification, and action recognition. 
Scene graph generation parses images into graphs representing objects and their relations, often combined with object detection~\cite{xu2017scene,yang2018graph}. 
Point clouds from LiDAR scans are effectively classified and segmented using GCNs~\cite{landrieu2018large,edgeconv}. 
Additionally, GCNs process human joints' graphs for activity recognition~\cite{yan2018spatial,cheng2020skeleton,liu2020disentangling}. Recent work even suggests using GNNs as backbone feature extractors for visual tasks, representing images as graphs~\cite{han2022vision}.

Graph-driven models have also been proposed for 6D pose estimation to better infuse structure information~\cite{chen2021fs, zhang2021keypoint, zhou2021pr}. They utilize orientation-aware autoencoders with 3D graph convolutions, which are robust to point shift and object size variations and construct k-NN graphs, extracting geometry-aware inter-modality correlations. Lian~\etal\cite{lian2023checkerpose} adopts GNNs to model interactions among 3D keypoints, aiming for 2D correspondences in input images. 

Inspired by these approaches, we integrate GCNs into the decoder's architecture to enhance semantic connections between keypoints and share information among structurally close ones.

\section{Method}
\begin{figure*}[t]
	\centering
	\includegraphics[width=0.98\textwidth]{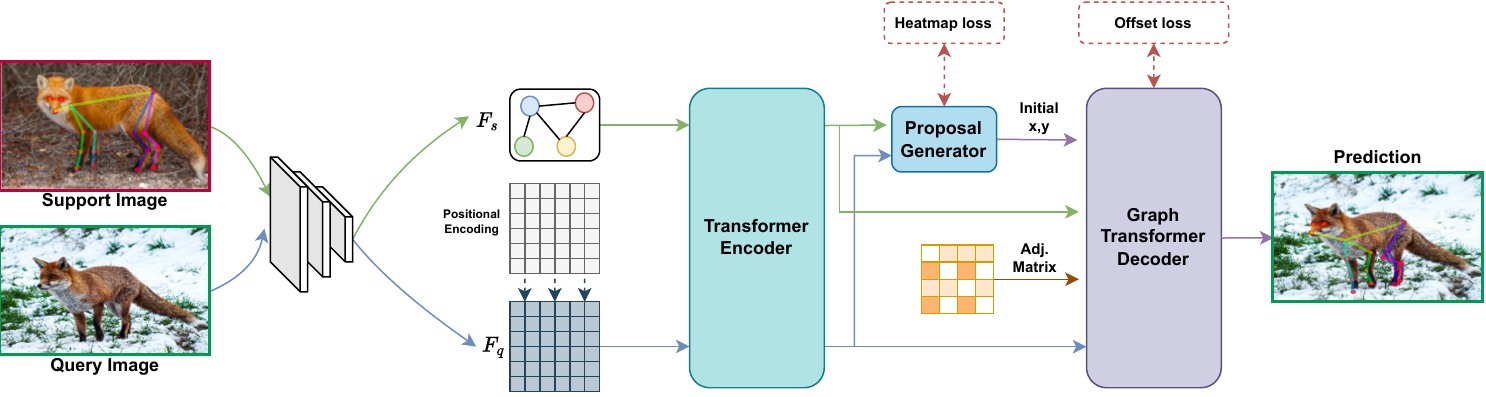}
	\caption{\textbf{Architecture Overview.} Our approach utilizes a pre-trained backbone to extract image features, followed by a transformer encoder that refines these features through self-attention. A similarity proposal generator is employed alongside a graph transformer decoder, enhancing keypoint localization accuracy with a focus on graph-oriented decoding.}
	\label{fig:intro}
\end{figure*}
In the following section, we describe our graph-based approach that takes advantage of the strong graph structure available in the data. The complete architecture of our method is illustrated in Figure~\ref{fig:intro}. 
Our approach utilizes a pre-trained backbone to extract support keypoints and query image features, followed by a transformer encoder that refines these features through self-attention. A similarity proposal generator is employed for initial coordinates localization. Finally, a transformer decoder predicts the keypoints' locations. More details on the different components are in the supplementary.

\subsection{A Graph-Based Approach}
\begin{figure}
  \centering
  \begin{tabular}{ccc}
    \includegraphics[width=0.23\textwidth]{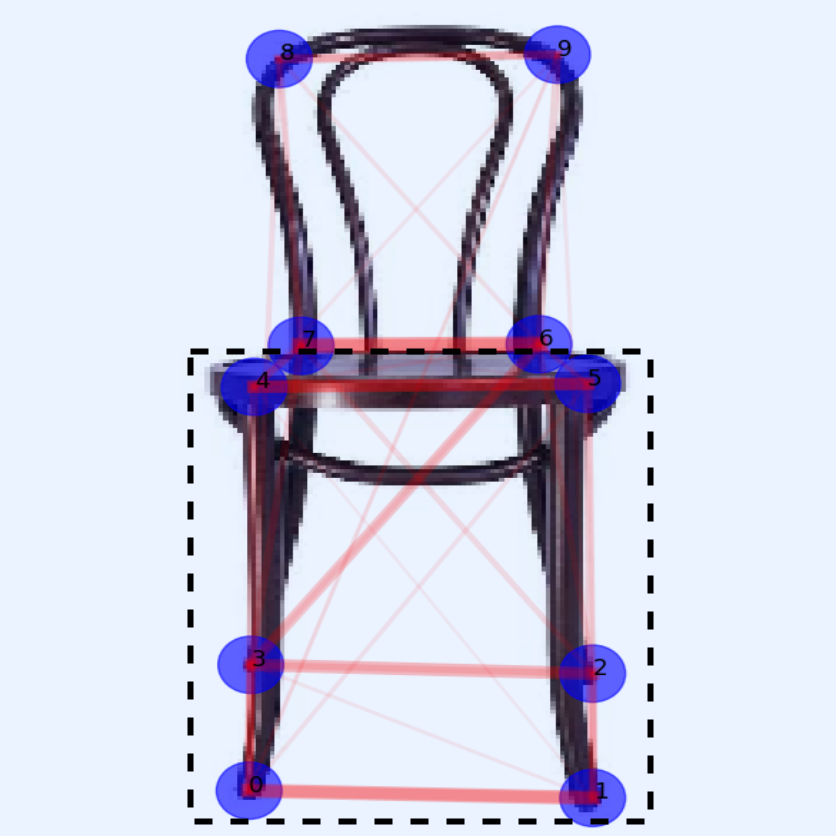} & 
    \includegraphics[width=0.23\textwidth]{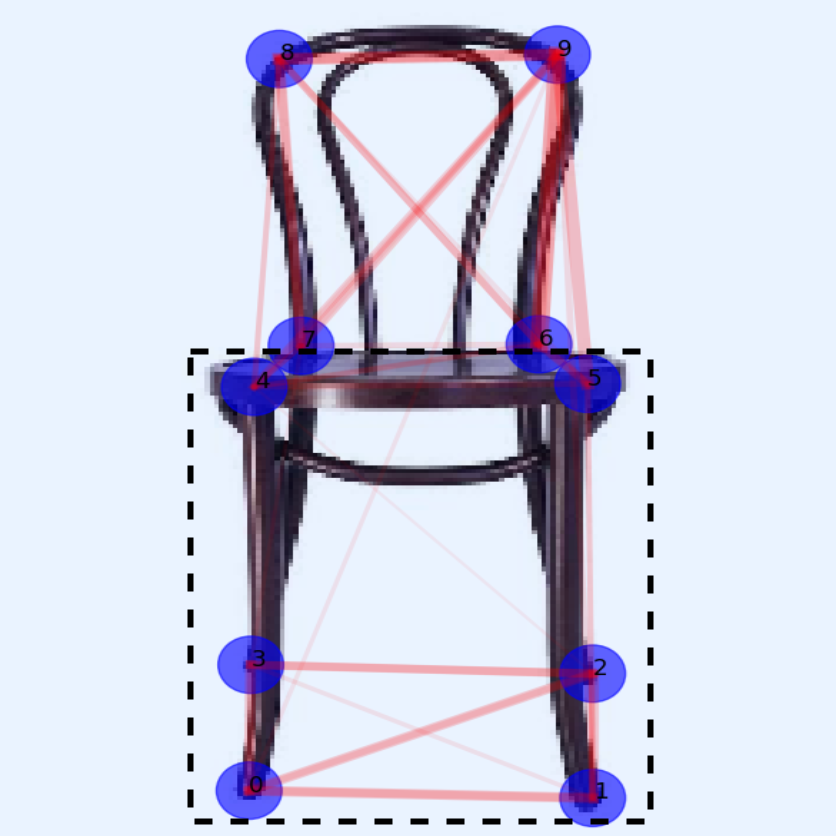} &
    \includegraphics[width=0.23\textwidth]{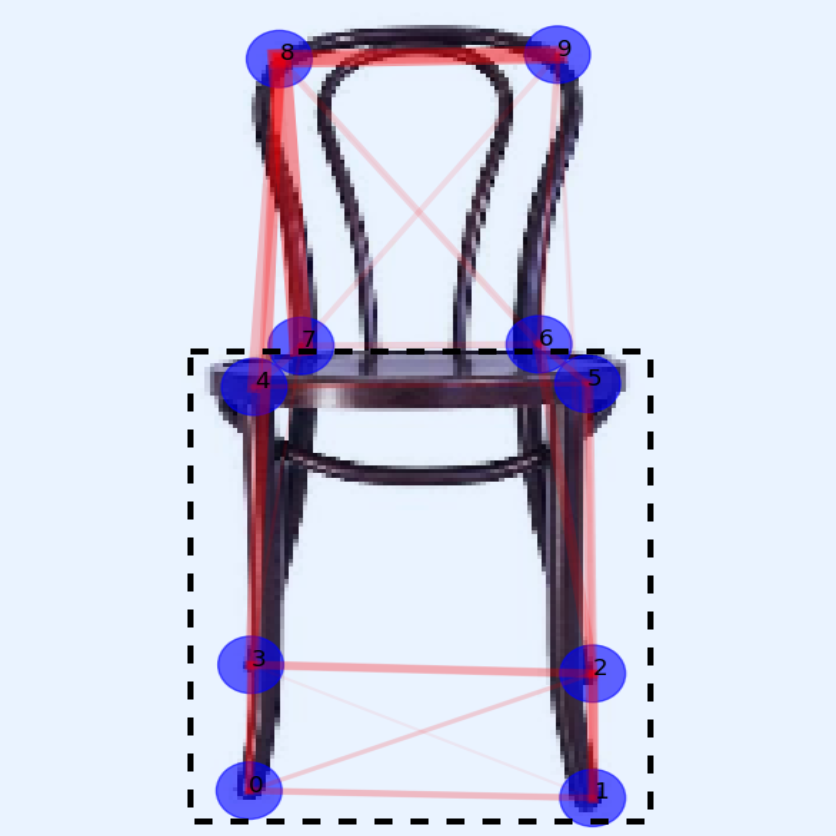} \\
    Multi-Category & Single-Category & Multi-Category \\
    Base Model & Base Model & Graph Model

    \end{tabular}
  \caption{\textbf{Self-Attention Map Visualization}. Comparing self-attention in decoders of three models: (a) CapeFormer-T trained on various object categories, (b) CapeFormer-T trained only on furniture objects, and (c) GraphCape trained on various object categories using a graph structure. Observe the edges between the legs and base of the chair. Notably, our graph-based method, despite being trained on multiple categories, exhibits similar attention patterns to models trained on single categories.
  }
  \label{fig:self_attn_map}
\end{figure}

The core idea of our work is in treating the input keypoints as a graph, to take advantage of the geometrical structure encoded in the graph connectivity. 
Our key insight is in recognizing that self-attention, a mechanism that helps models focus on relevant information, can be thought of as a GCN with an input-dependent learnable adjacency matrix. When we are dealing with pose estimation for a single category, this mechanism is sufficient for learning the relationships between keypoints and integrating a learned structure into the model.
However, for the new task of CAPE, where the model needs to work with object categories it has never seen before, it is beneficial to explicitly consider the semantic connections between keypoints. 
As GNN promotes information sharing between connected keypoints, location data is shared among structurally connected keypoints, which are typically close. Thus, the graph design aids in structure preservation and occlusion handling. 
Moreover, we believe that utilizing the graph structure during training helps in breaking feature symmetry, as it enforces similar features based on structure (e.g., right/left) rather than solely on semantics.
Figure~\ref{fig:self_attn_map} illustrates this hypothesis qualitatively. Notably, our graph-based method, despite being trained on multiple categories, exhibits similar attention patterns to models trained on single categories.

\begin{figure}[t]
	\centering
	\includegraphics[width=0.95\textwidth]{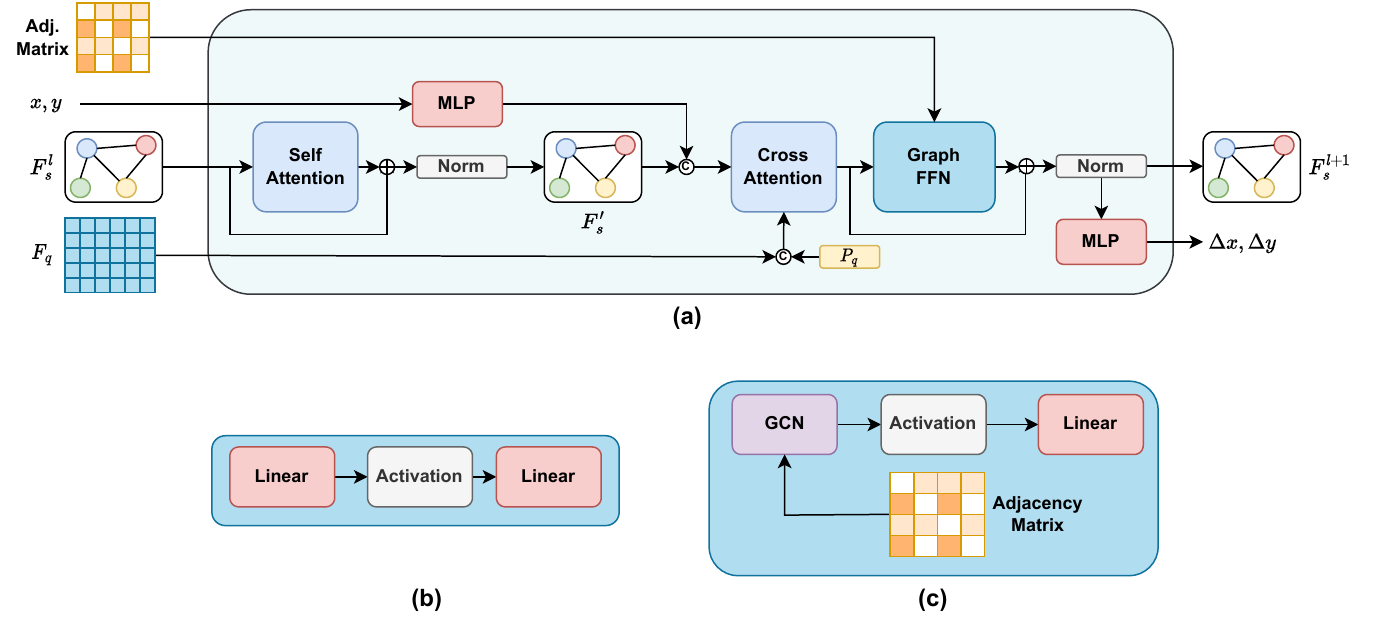}
	\caption{\textbf{Graph FFN.} The Transformer decoder is based on the original CapeFormer design, changing the feed-forward network from a simple MLP to a graph-based network.
 \textbf{(a)} A scheme of the transformer decoder which includes self-attention, cross-attention, and a feed-forward network. Self-attention encourages adaptive interactions among support keypoints, while cross-attention extracts localization information. 
 \textbf{(b)} Previous FFN consisted of an MLP with 2 layers. 
 \textbf{(c)} Our graph FFN includes a GCN layer and subsequent linear layers that enhance keypoint features and promote information exchange among known connected keypoints.
 }
	\label{fig:graph_decoder}
\end{figure}

We implemented this prior into the transformer decoder. Specifically, our design is based on CapeFormer, changing the decoder's feed-forward network from a simple MLP to a GCN-based module. To address the potential problem of excessive smoothing often observed in deep GCNs~\cite{li2018deeper, oono2019graph}, which can lead to a reduction in the distinctiveness of node characteristics and consequently a decline in performance, we introduce a linear layer for each node following the GCN layer. Figure~\ref{fig:graph_decoder} illustrates our new design.

The decoder has three main components: self-attention, cross-attention, and a feed-forward network. 
Self-attention allows for adaptive interactions among support keypoints, transforming the input feature $F_s^l$ into $F_s'$. 
Cross-attention augments support keypoints' representation by extracting localization information from query feature patches, enhancing attention at proposed locations. This involves concatenating the keypoint localization embedding with keypoint features. Cross-attention inputs are $F_s'$ (with the localization embedding) as queries, patch features $F_q$ (with positional encoding) as keys, and $F_q$ as values, resulting in transformed features $F_s''$.
Lastly, we employ a graph-based feed-forward network to process the output keypoint features as a graph.  This layer further concentrates the keypoint features, facilitating the exchange of information between neighboring keypoints in the graph.

The output value for an input $F_s'' \in \mathbb{R}^{C_{in} \times K}$ can be written as:
\begin{equation}
\widetilde{F_s''} =\sigma_{act} (W_{adj}F_s''\widetilde{A} + W_{self}F_s'')
\end{equation}
\begin{equation}
F_s^{l+1} = F_s'' + W_{linear}\widetilde{F_s''}
\end{equation}
Where $W_i \in \mathbb{R}^{C_{out} \times C_{in}}$ is a learnable parameter matrix, $\sigma_{act}$ is an activation function (ReLU), and $\widetilde{A}\in \mathbb{R}^{K \times K}$ is the symmetrically normalized form of the adjacency matrix $A \in [0,1]^{K \times K}$. $A$ is a binary matrix, defined as $a_{ij} = 1$ for node $v_j$ that is connected to node $v_i$ and $0$ elsewhere. 
While the adjacency matrix is fixed for each category, during training some nodes may not be visible in the support image. Thus, we also mask the adjacency matrix before normalization, to make sure only meaningful information is propagated between nodes.

The output features $F_s^{l+1}$ are used to update the normalized keypoints locations from $P^l$ to $P^{l+1}$. We follow~\cite{Shi_2023_CVPR} and use the following update function:
\begin{equation}
 P^{l+1} = \sigma(\sigma^{-1}(P^{l}) + MLP (F_s^{l+1})) 
\end{equation}
where $\sigma$ and $\sigma^{-1}$ are the sigmoid and its inverse function.
The keypoints' positions from the last decoder layer are used as the final keypoints prediction.

\subsection{Training Scheme}
Following CapeFormer, we use two supervision signals: a heatmap loss and an offset loss. The heatmap loss supervises the proposal generator, constraining the shape of the similarity maps, and facilitating the learning of meaningful representations, while the offset loss supervises the localization output:
\begin{equation}
\mathcal{L}_{heatmap} = \frac{1}{K \cdot H \cdot W}\sum_{i=1}^K \lVert \sigma(M_i)-H_i \rVert
\end{equation}
\begin{equation}
\mathcal{L}_{offset} = \frac{1}{L}\sum_{i=1}^L\sum_{i=1}^K |P_i^l - \hat{P_i}|
\end{equation}
where $M_i$ denotes the output similarity heatmap from the proposal generator, $\sigma$ is the sigmoid function, $H_i$ the ground-truth heatmap, $P_i$ the output location from the each layer and $ \hat{P_i}$ the ground-truth location.
The overall loss term is:
\begin{equation}
\mathcal{L} = \lambda_{heatmap} \cdot \mathcal{L}_{heatmap} + \mathcal{L}_{offset}
\end{equation}

\subsection{Encoding Structure: Positional Encoding VS Graph Network}
\begin{table}[t]
\centering
\caption{\textbf{Keypoint Positional Encoding:} Comparison of results using CapeFormer with the original dataset and using permuted keypoints. Keypoint Positional Encoding (PE) adds an implicit structure bias according to the fixed order of keypoints in the dataset. Graph-FFN adds an explicit structure prior of any object's shape.}
\begin{tabular}{cccc}
\toprule
 & Original Data & \quad \quad \quad & Shuffled Keypoints \\ 
\midrule
With PE & 89.26 &&  64.20 \textbf{(-25.06\%)} \\
Without PE & 85.24 &&  85.24 \textbf{(0\%)}\\
Without PE + Graph-FFN & 87.71 && 87.71 \textbf{(0\%)}\\
\end{tabular}
\label{tab:pos_encoding}
\end{table} 
CapeFormer introduced a keypoint positional encoding (PE), termed the "Support Keypoint Identifier". This encoding is generated via sinusoidal encoding of the keypoints' order.
We analyzed this feature thoroughly and present the results in Table~\ref{tab:pos_encoding}.
As shown, this encoding improves performance by a significant 4\% PCK (defined later), at the cost of introducing a dependency on the order of keypoints. 
This became evident when we evaluated a model trained with this positional encoding on data of keypoints in permuted order. Using PE and shuffling the keypoints results in a notable decrease of approximately 25\% in PCK, indicating that the model had become highly specialized for a specific keypoint format. This sharp decrease is not present when the positional encoding is not used.
Although the dataset contains diverse object categories, some of them share a common keypoint ordering.
This indicates that using this PE adds structure information, learned from known categories in the training set.

As we require invariance to keypoint permutations, graph structure is an obvious choice. Our graph-based approach serves as an alternative, boosting performance while being agnostic to permutations in the order of nodes.
\section{Experiments}
Following previous CAPE works, we use the MP-100 dataset~\cite{xu2022pose} for training and evaluation, comprising samples drawn from existing category-specific pose estimation datasets.
The MP-100 dataset encompasses a collection of over 20,000 images, spanning 100 distinct categories, with varying keypoint numbers up to 68 across different categories.
To ensure categories used for evaluation remain unseen during the training process, the samples are divided into five different mutually exclusive splits. 
The dataset includes partial skeleton annotations in various formats, including differences in the indexing of keypoints (some start from zero while others from one). We adopted a unified format with comprehensive skeleton definitions for all categories (skeleton annotations are shared among categories). This process involved cross-referencing the original datasets and, where necessary, conducting manual annotations.

To quantify our model's performance, we employ the Probability of Correct Keypoint (PCK) metric~\cite{yang2012articulated}, with a PCK threshold set at 0.2, following previous conventions~\cite{xu2022pose, Shi_2023_CVPR}.

\subsubsection{CapeFormer-T.} 
In the following sections, we compare our model to previous works along with an updated version of the previous state-of-the-art method which we call CapeFormer-T. 
For this baseline, we replaced the original ResNet-50~\cite{he2016deep} backbone with a stronger transformer-based SwinV2-T~\cite{liu2022swin}. After testing various configurations, including multi-scale and single-scale feature maps, we found that applying bilinear upsampling on the final feature layer achieved comparable results while maintaining simplicity. To capitalize on the improved feature quality we extract keypoint features using a Gaussian kernel mask with lower variance.
We believe that CAPE should not enforce a specific keypoint order. Therefore, following the discussion in the previous section, we removed keypoint positional encoding.
These simple adjustments led to an improvement of 0.94\% over the original CapeFormer. We provide a short description of the different components of CapeFormer in the supplementary.

It is important to note that the difference between CapeFormer-T and our method GraphCape, is the graph-based decoder design, which utilizes the graph structure during training and inference. Thus any performance difference between these models is a result of using our graph-oriented architecture.

\subsection{Implementation Details}
For a fair comparison, network parameters, training parameters, data augmentations and data pre-processing are kept the same as CapeFormer~\cite{Shi_2023_CVPR}:
Both the encoder and decoder have three layers and $\lambda_{heatmap}$ is set to $2.0$. 
The model is built upon MMPose framework~\cite{mmpose2020}, trained using Adam optimizer for 200 epochs with a batch size of 16, the learning rate is $10^{-5}$, and decays by 10× on the 160th and 180th epoch. More design choices and evaluations are in the supplementary.

\subsection{Qualitative Results}
\begin{figure*}
\setlength{\fboxsep}{0pt}
\setlength{\fboxrule}{1.5pt}
  \centering
 \scriptsize
  \begin{tabular}{c ccccc}
     Support & GT & POMNet & CapeFromer & CapeFormer-T & \textbf{GraphCape} \\
     
    \fcolorbox{purple}{white}{\includegraphics[width=0.15\textwidth]{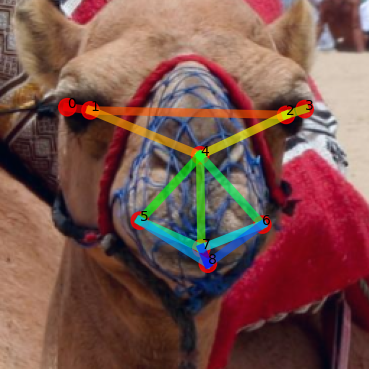}} & 
    \fcolorbox{ForestGreen}{white}{\includegraphics[width=0.15\textwidth]{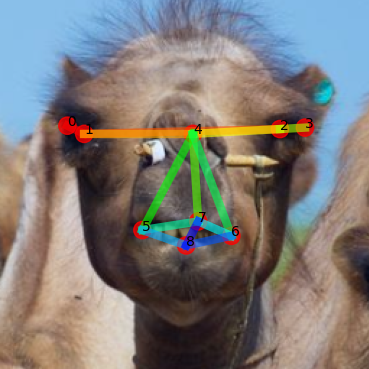}} & 
    \fcolorbox{ForestGreen}{white}{\includegraphics[width=0.15\textwidth]{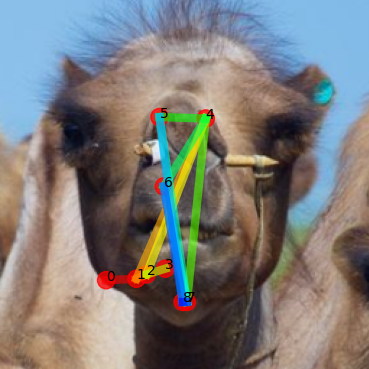}} & 
    \fcolorbox{ForestGreen}{white}{\includegraphics[width=0.15\textwidth]{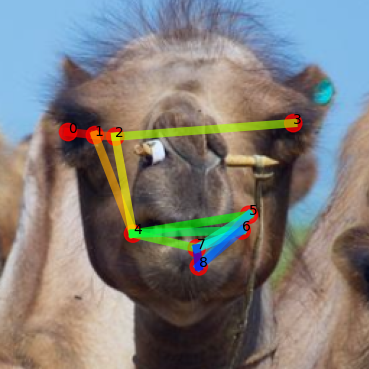}} &
    \fcolorbox{ForestGreen}{white}{\includegraphics[width=0.15\textwidth]{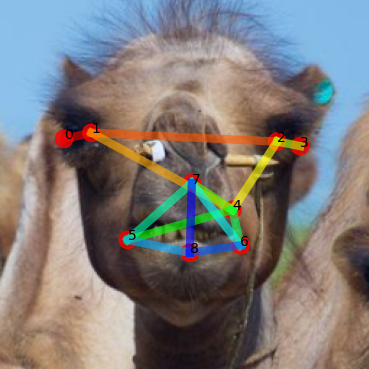}} &
    \fcolorbox{ForestGreen}{white}{\includegraphics[width=0.15\textwidth]{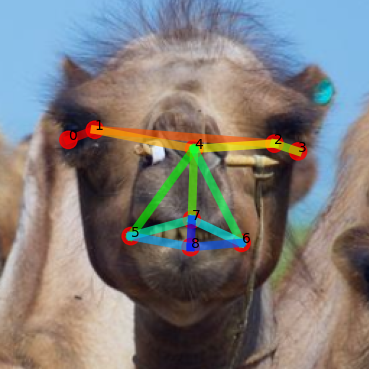}} \\
    \fcolorbox{purple}{white}{\includegraphics[width=0.15\textwidth]{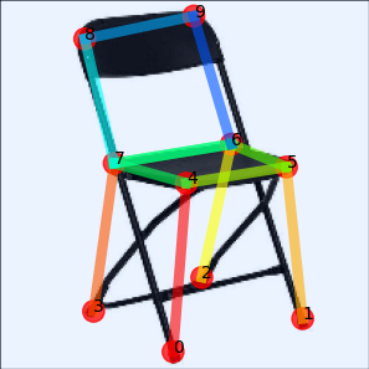}} & 
    \fcolorbox{ForestGreen}{white}{\includegraphics[width=0.15\textwidth]{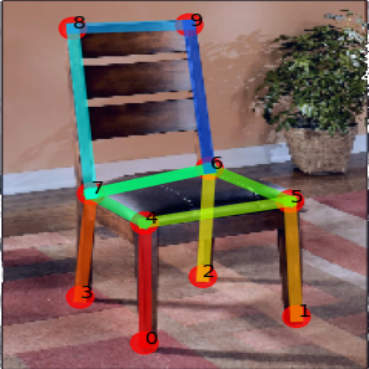}} & 
    \fcolorbox{ForestGreen}{white}{\includegraphics[width=0.15\textwidth]{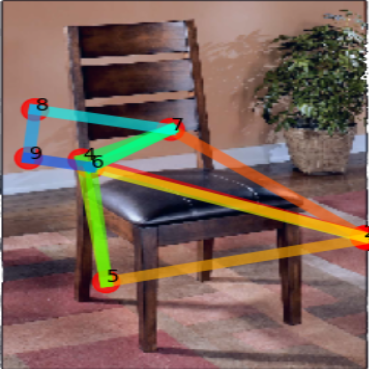}} &
    \fcolorbox{ForestGreen}{white}{\includegraphics[width=0.15\textwidth]{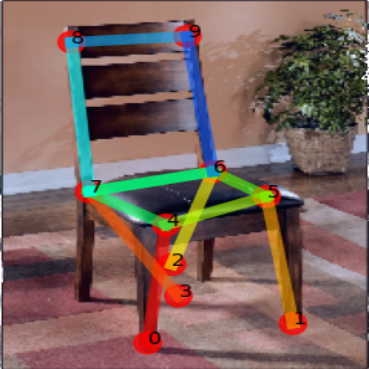}} & 
    \fcolorbox{ForestGreen}{white}{\includegraphics[width=0.15\textwidth]{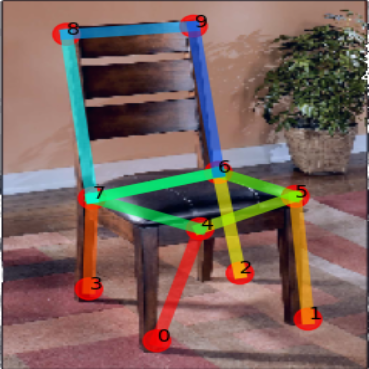}}  &
    \fcolorbox{ForestGreen}{white}{\includegraphics[width=0.15\textwidth]{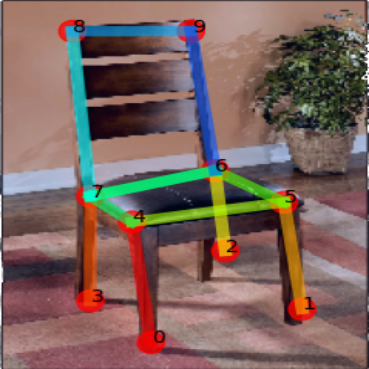}} \\
    
    \fcolorbox{purple}{white}{\includegraphics[width=0.15\textwidth]{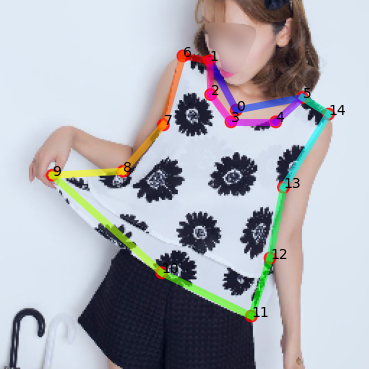}} & 
    \fcolorbox{ForestGreen}{white}{\includegraphics[width=0.15\textwidth]{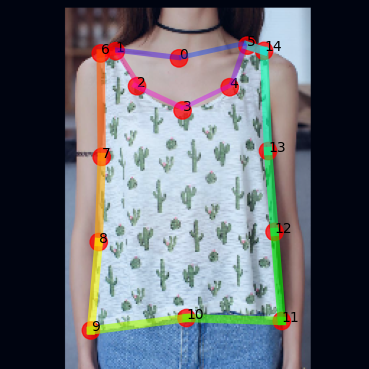}} & 
    \fcolorbox{ForestGreen}{white}{\includegraphics[width=0.15\textwidth]{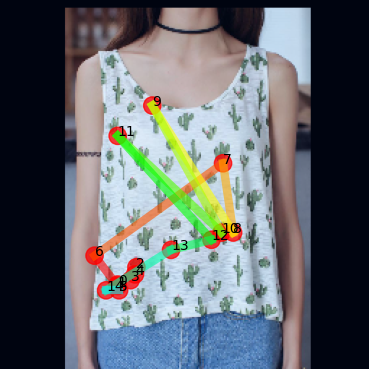}} & 
    \fcolorbox{ForestGreen}{white}{\includegraphics[width=0.15\textwidth]{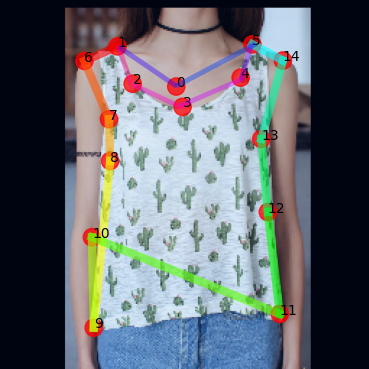}} &
    \fcolorbox{ForestGreen}{white}{\includegraphics[width=0.15\textwidth]{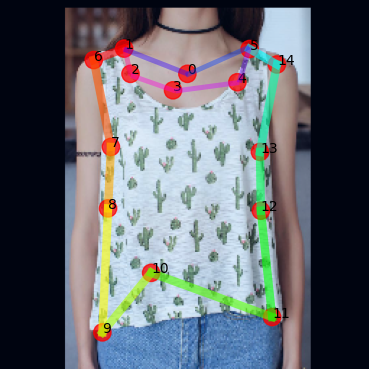}} &
    \fcolorbox{ForestGreen}{white}{\includegraphics[width=0.15\textwidth]{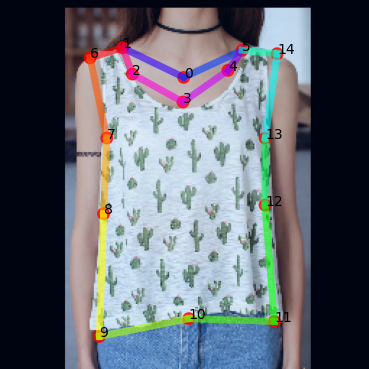}} \\

    \end{tabular}
  \caption{\textbf{Qualitative Results.} We visualize the keypoint predictions under a 1-shot setting. The left column denotes the support image with its corresponding skeleton. The second column is the ground-truth query keypoints. The following columns are results from POMNet, CapeFormer, CapeFormer-T, and our method.
  }
  \label{fig:qualitative}
\end{figure*}

We show a qualitative comparison between our method, CapeFormer-T, and previous CAPE methods: CapeFormer~\cite{Shi_2023_CVPR} and POMNet\footnote{We used POMNet official code to train and visualize results, as they don't provide pre-trained models. Quantitative results are taken from the published paper.}~\cite{xu2022pose} in Figure~\ref{fig:qualitative}.
As can be seen, the structure information incorporated in our method serves as a strong prior for localizing keypoints, helping in breaking symmetry and creating structure consistency between the keypoints. The first row of the figure shows an example of symmetry that confuses prior methods but not ours, while the third row shows our method's superiority in preserving structure consistency across keypoints. More examples are in the supplementary.

\subsection{Quantitative Results}
\begin{table*}[t]
\centering
\caption{
\textbf{MP-100 Results.} PCK performance under 1-shot and 5-shot settings. In both settings, our approach consistently outperforms other methods, across all splits.}
\begin{tabular}{cc c|ccccc | c}

\toprule
&& \textbf{Model} & Split 1 & Split 2 & Split 3 & Split 4 & Split 5 & Avg \\
\midrule
\multirow{7}{*}{\textbf{1-Shot}} && ProtoNet~\cite{snell2017prototypical} & 46.05 & 40.84 & 49.13 & 43.34 & 44.54 & 44.78 \\
&& MAML~\cite{finn2017model} & 68.14 & 54.72 & 64.19 & 63.24 & 57.20 & 61.50 \\
&& Fine-tuned~\cite{nakamura2019revisiting} & 70.60 & 57.04 & 66.06 & 65.00 & 59.20 & 63.58 \\
&& POMNet~\cite{xu2022pose} & 84.23 & 78.25 & 78.17 & 78.68 & 79.17 & 79.70 \\
&& CapeFormer~\cite{Shi_2023_CVPR} &  89.45 & 84.88 & 83.59 & 83.53 & 85.09 & 85.31 \\
\cmidrule(lr){2-9}

&& CapeFormer-T & 89.48 & 86.69 & 85.31 & 84.79 & 84.97 & 86.25  \\
&& \textbf{GraphCape} & \textbf{91.19} & \textbf{87.81} & \textbf{85.68} & \textbf{85.87} & \textbf{85.61} & \textbf{87.23} \\

\midrule

\multirow{7}{*}{\textbf{5-Shot}} && ProtoNet~\cite{snell2017prototypical} & 60.31 & 53.51 & 61.92 & 58.44 & 58.61 & 58.56 \\
&& MAML~\cite{finn2017model} & 70.03 & 55.98 & 63.21 & 64.79 & 58.47 & 62.50 \\
&& Fine-tuned~\cite{nakamura2019revisiting} & 71.67 & 57.84 & 66.76 & 66.53 & 60.24 & 64.61 \\
&& POMNet~\cite{xu2022pose} & 84.72 & 79.61 & 78.00 & 80.38 & 80.85 & 80.71 \\
&& CapeFormer~\cite{Shi_2023_CVPR} & 91.94 & 88.92 & 89.40 & 88.01 & 88.25 & 89.30 \\
\cmidrule(lr){2-9}

&& CapeFormer-T & 94.04 & 91.20 & 89.19 & \textbf{90.63} & 89.45 & 90.90 \\
&& \textbf{GraphCape} & \textbf{94.24} & \textbf{91.32} & \textbf{90.15} & 90.37 & \textbf{89.73} & \textbf{91.16} \\


\bottomrule
\end{tabular}
\label{Tab:mp100}
\end{table*} 
We compare our method with CapeFormer-T, previous CAPE methods CapeFormer~\cite{Shi_2023_CVPR} and POMNet~\cite{xu2022pose}, and three baselines:
ProtoNet~\cite{snell2017prototypical}, MAML~\cite{finn2017model}, and Fine-tuned~\cite{nakamura2019revisiting}. Further details on these model's evaluation can be found in~\cite{xu2022pose}.
We report results on the MP-100 dataset under 1-shot and 5-shot settings in Table~\ref{Tab:mp100}. As can be seen, our graph-based method consistently improves performance over CapeFormer-T on the different dataset splits, on average by 0.98\% under the 1-shot setting and 0.26\% under the 5-shot setting, achieving new state-of-the-art results for both settings.

In the supplementary, we also show better scalability of our design. Similar to DETR-based models, employing a larger backbone improves performance and our graph-based design also enhances performance using larger backbones. Furthermore, we show that our graph model scales better with the number of decoder layers compared to CapeFormer-T.

\subsection{Ablation Study}
We conducted several ablation studies on the MP-100 dataset. 
We first show quantitative examples using out-of-distribution images. Then, we display the contribution of the geometrical structure prior by evaluating our model using different skeleton relations. Then, we demonstrate the advantages of using graph structure in handling occlusions by evaluating the performance using masked inputs. Lastly, we test our model's performance under a cross-category setting.
We perform all ablation experiments on the test set of MP-100 split1 under the 1-shot setting following~\cite{Shi_2023_CVPR, xu2022pose}.
In the supplementary, we also show our model's performance with various pre-trained backbones, aligning with DETER-based results~\cite{lin2023detr} regarding the preferred backbone. 

\subsubsection{Out-of-Distribution Performance.}
\begin{figure}
\setlength{\fboxsep}{0pt}
\setlength{\fboxrule}{1.5pt}
\renewcommand{\arraystretch}{1}
  \centering
  \begin{tabular}{ccccc}
  
    \fcolorbox{purple}{white}{\includegraphics[width=0.2\textwidth]{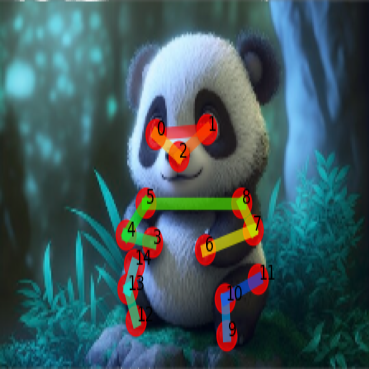}} &
    \fcolorbox{purple}{white}{\includegraphics[width=0.2\textwidth]{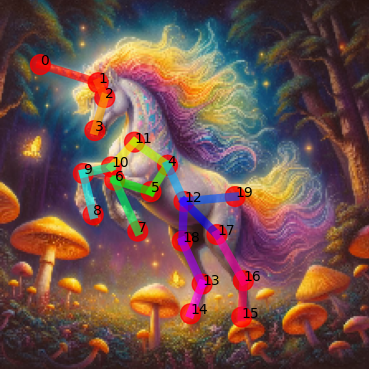}} & &
    \fcolorbox{purple}{white}{\includegraphics[width=0.2\textwidth]{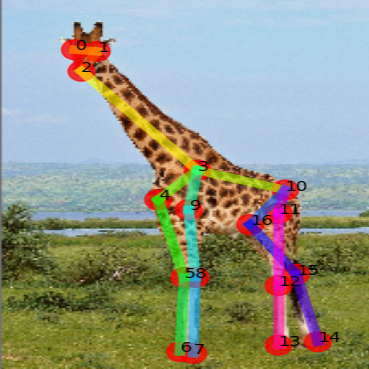}} &
    \fcolorbox{purple}{white}{\includegraphics[width=0.2\textwidth]{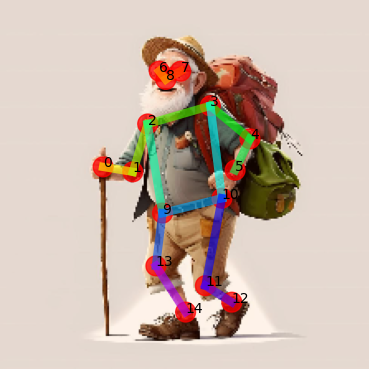}} \\
    \fcolorbox{ForestGreen}{white}{\includegraphics[width=0.2\textwidth]{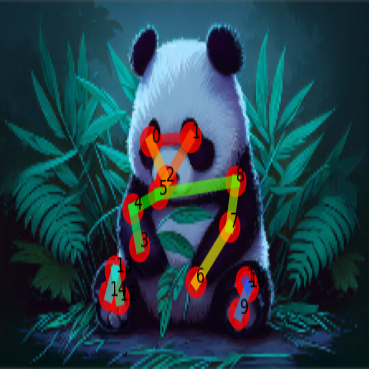}} &
    \fcolorbox{ForestGreen}{white}{\includegraphics[width=0.2\textwidth]{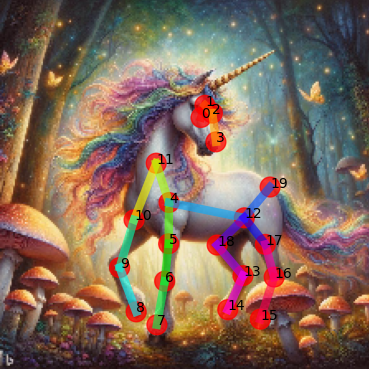}} &  &
    \fcolorbox{ForestGreen}{white}{\includegraphics[width=0.2\textwidth]{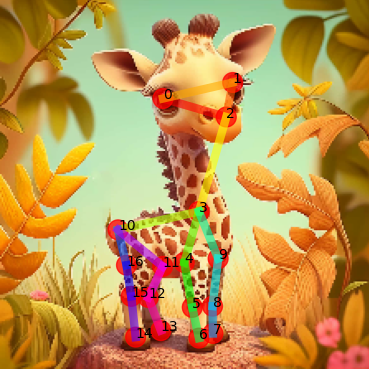}} &
    \fcolorbox{ForestGreen}{white}{\includegraphics[width=0.2\textwidth]{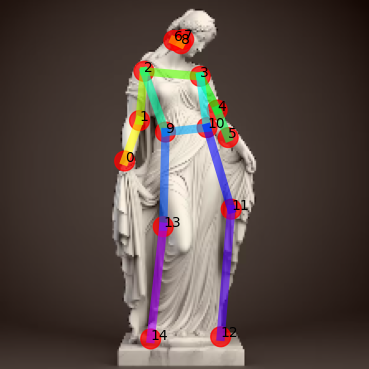}} \\
    \multicolumn{2}{c}{\textbf{(a)}} & & \multicolumn{2}{c}{\textbf{(b)}} \\

    \end{tabular}
  \caption{\textbf{Out-of-Distribution.} Qualitative results using OOD samples, \textcolor{purple}{top} is the support image and \textcolor{ForestGreen}{bottom} the query. \textbf{(a)}: Support and query are from the same OOD domain. \textbf{(b)}: Support and query images are from different domains.
  }
  \label{fig:ood}
\end{figure}

To assess the robustness of our model, we evaluate our network using images from different domains. Results are shown in Figure~\ref{fig:ood}. Our model, which was trained on real images only, demonstrates its adaptability and effectiveness across varying data sources such as cartoons and imaginary animals, created using a text-to-image diffusion model. 
Furthermore, our model performs satisfactorily even when the support and query images are from different domains.
However, when the images are from vastly different categories from the ones seen during training, the performance degrades. 
Yet, this is common to all CAPE methods and doesn’t relate to using graphs.

\subsubsection{The Contribution of Graph Structure.} 
To assess the contribution of the graph structure prior, we evaluate our method using different graph inputs. Results are shown in Table~\ref{tab:graphs}.
Training and evaluating the graph model without node's connectivity, the model is equivalent to CapeFormer-T, thus performing the same.
Moreover, training the model on fully-connected graphs results in a similar performance to CapeFormer-T as mathematically it is equivalent to adding a bias vector to all nodes, which is normalized and canceled. And again, performance is similar to CapeFormer-T (intuitively, as all graph inputs are the same, no new information is provided).
When providing random graphs the model's performance degrades as wrong connectivity involves inferior semantic features for localization.
This shows a limitation of using graphs, as the wrong skeleton data may impair performance. 
Lastly, we show in the supplementary that different graph definitions may result in different performances. However, we noticed empirically that most straightforward skeletal definitions produce similar results. Furthermore, as we didn’t optimize the dataset skeletons, we believe that further quantitative improvements can be made using optimal skeletal relations. Thus, we defer the exploration of the optimal skeletons to future research.
\begin{table}[t]
\centering
\caption{
\textbf{Different Graph Connectivity:} We evaluate our graph-oriented network, training and evaluating with different types of graph definitions.
}
\small
\begin{tabular}{cc}
\toprule
Graph Structure & PCK$_{0.2}$\\ 
\midrule
No Connectivity & 89.48 \\
Fully Connected & 89.35 \\
Random Graphs (each instance) & 81.49 \\
Random Graphs (each category) & 88.10 \\
Pose Graph & \textbf{91.19}\\
\end{tabular}
\label{tab:graphs}
\end{table} 

\subsubsection{Masking Support/Query Images.}
\begin{figure}
\centering
\begin{subfigure}[b]{0.7\textwidth}
\centering
    \includegraphics[width=0.95\textwidth]{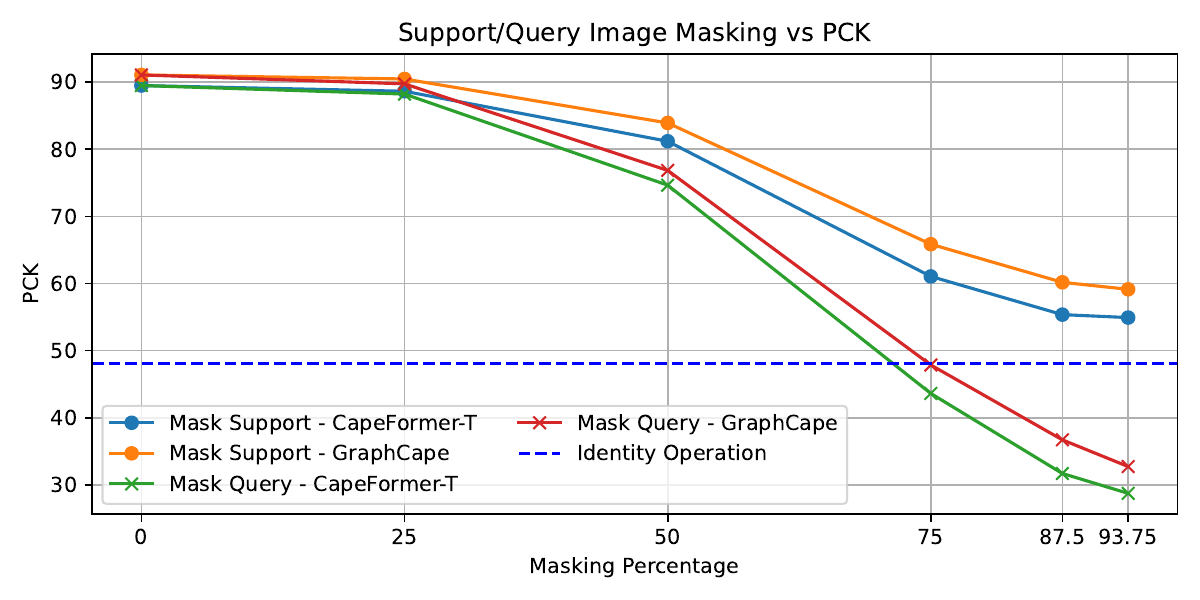}
    \caption{}\label{fig:mask_a}
\end{subfigure} \hfill
\begin{subfigure}[b]{0.85\textwidth}
\setlength{\fboxsep}{0pt}
\setlength{\fboxrule}{1.5pt}
\setlength{\tabcolsep}{1pt}
    \centering
        \begin{tabular}{ccccc}
        \fcolorbox{purple}{white}{\includegraphics[width=0.23\textwidth]{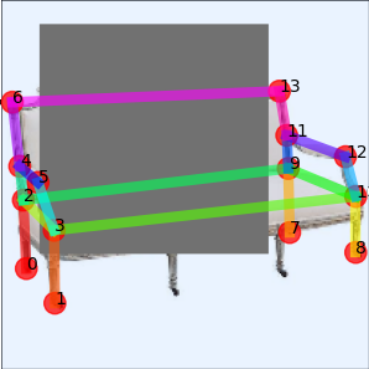}} & &
        \fcolorbox{ForestGreen}{white}{\includegraphics[width=0.23\textwidth]{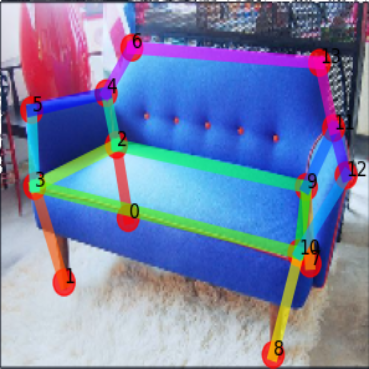}} & 
        \fcolorbox{ForestGreen}{white}{\includegraphics[width=0.23\textwidth]{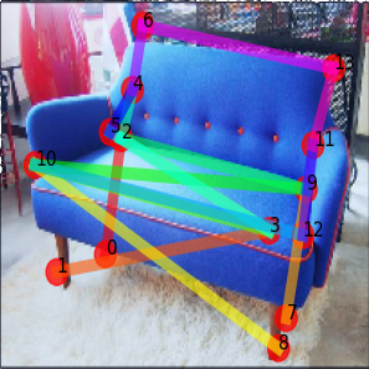}} &
        \fcolorbox{ForestGreen}{white}{\includegraphics[width=0.23\textwidth]{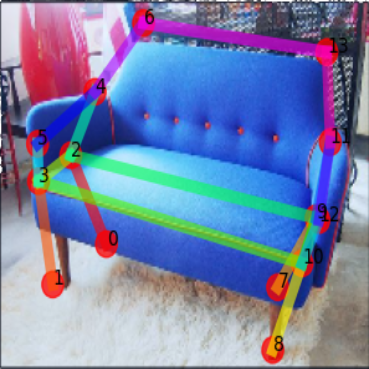}} \\

        \fcolorbox{purple}{white}{\includegraphics[width=0.23\textwidth]{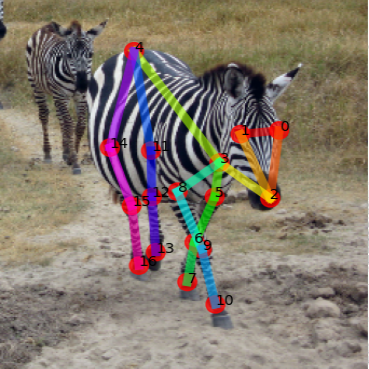}} & &
        \fcolorbox{ForestGreen}{white}{\includegraphics[width=0.23\textwidth]{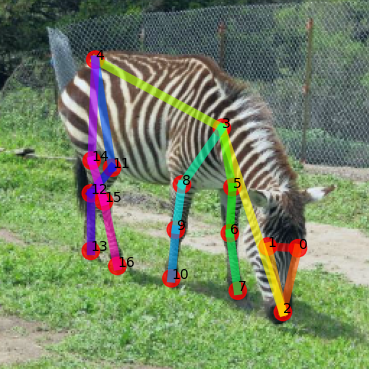}} & 
        \fcolorbox{ForestGreen}{white}{\includegraphics[width=0.23\textwidth]{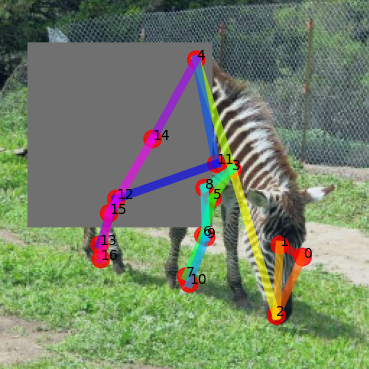}} &
        \fcolorbox{ForestGreen}{white}{\includegraphics[width=0.23\textwidth]{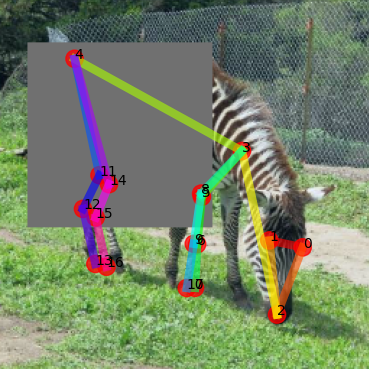}} \\
        
        Support && GT & CapeFormer-T & GraphCape
    \end{tabular}
    \caption{}\label{fig:mask_b}
\end{subfigure}
\caption{\textbf{Masking.} 
\textbf{(a)} Quantitative results when masking the support/query images. 
Our method consistently surpasses CapeFormer-T, leveraging cues provided by the graph structure to overcome information gaps in the support/query images.
\textbf{(b)} Qualitative results when masking the support (\textbf{top}) and query (\textbf{bottom}) image using CapeFormer-T and our graph-based model. Our model better handles occlusions by preserving structure and breaking symmetry.
}
\label{fig:mask}
\end{figure}

To highlight the contribution of graph information in handling occlusions, we applied partial masking to the support/query image before executing our algorithm. It is important to note that the difference between the two compared models is our suggested graph-based decoder.
Quantitative analysis in Figure~\ref{fig:mask_a} shows our method consistently outperforming CapeFormer-T. The dashed horizontal line represents the identity operation which outputs the support keypoint's locations, preserving structure.
Notably, our model can predict keypoints even when a significant portion of the support image is masked. This suggests that the model has learned which keypoints are relevant and matches them based on structure. 
For instance, in Figure~\ref{fig:mask_b} (top), the model accurately predicts most keypoints. In comparison, CapeFormer-T is confused by symmetry, breaking the structure of the sofa.
However, when large portions of the query image are masked, the model's performance rapidly declines, although it retains structure, as evident in Figure~\ref{fig:mask_b} (bottom). 
This experiment suggests that our model better handles occlusions, which is crucial for real-world applications.
More examples are in the supplementary.

\subsubsection{Cross-Category Correspondence.} 
\begin{figure}
\setlength{\fboxsep}{0pt}
\setlength{\fboxrule}{1.5pt}
\renewcommand{\arraystretch}{1}
  \centering
  \begin{tabular}{ccc}
    \fcolorbox{purple}{white}{\includegraphics[width=0.28\textwidth]{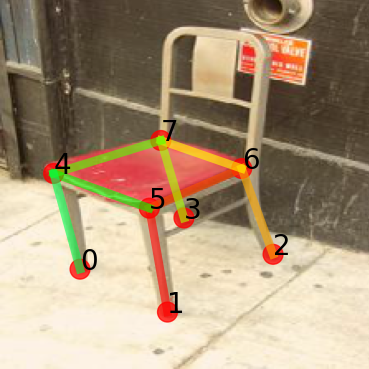}} & 
    \fcolorbox{ForestGreen}{white}{\includegraphics[width=0.28\textwidth]{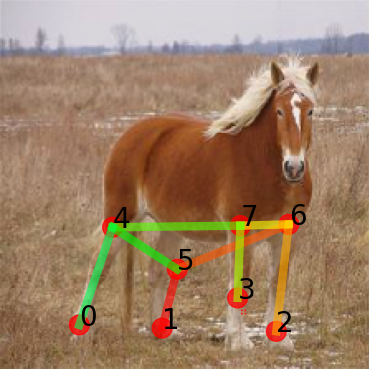}} & 
    \fcolorbox{ForestGreen}{white}{\includegraphics[width=0.28\textwidth]{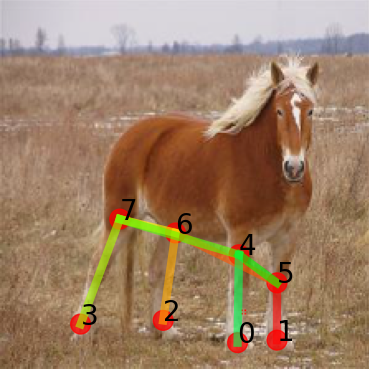}} \\
    Support & Capefromer-T & GraphCape \\

    \end{tabular}
  \caption{\textbf{Cross-Category Correspondence}. We test our model's performance when the support and query images are from vastly different categories. 
 The difference lies in matching keypoints. CapeFormer-T matches keypoints naively, based on their location in the image.
  In contrast, our model breaks the symmetry between the legs and correctly matches keypoints with the correct object 3D orientation.
  }
  \label{fig:cross}
\end{figure}

We test our model's performance when the support and query images are from vastly different categories. Results can be seen in Figure~\ref{fig:cross}. CapeFormer-T correctly finds correspondence between the legs of the chair and of the horse. However, it mistakenly matches keypoints according to their location in the image and not according to the orientation of the objects. Our model breaks the symmetry between the leg keypoints and correctly matches keypoints with the correct 3D orientation.

\section{Conclusion}
We present a novel approach for category-agnostic pose estimation (CAPE) by treating input keypoints as graphs, recognizing the importance of underlying geometrical structures within objects.
We implement this prior using a Graph-FFN, which captures and incorporates structural information by exploiting the relationships and dependencies between keypoints. This change significantly enhances the accuracy of keypoint localization.
In addition, we provide an updated version of the MP-100 dataset, which now includes skeleton annotations for all categories, further promoting research in CAPE.

Our experimental results demonstrate the superiority of our method over our updated version of the previous state-of-the-art approach, CapeFormer-T.
With improvements under both 1-shot and 5-shot settings, our method opens the door to more versatile and adaptable applications in computer vision.

\subsubsection{Acknowledgement.} Part of this research was supported by the Weinstein Institute and ISF grant 2132/23.
\clearpage
%
%
\bibliographystyle{splncs04}
\bibliography{egbib}
\end{document}